# TractGeoNet: A geometric deep learning framework for pointwise analysis of tract microstructure to predict language assessment performance

Yuqian Chen[a,b], Leo R. Zekelman[c,d], Chaoyi Zhang[b], Tengfei Xue[a,b], Yang Song[e], Nikos Makris[f], Yogesh Rathi[a,g], Alexandra J. Golby[c], Weidong Cai[b], Fan Zhang[a,h*], Lauren J. O'Donnell[a*]

[a] Department of Radiology, Brigham and Women's Hospital, Harvard Medical School, Boston, MA, USA

[b] School of Computer Science, The University of Sydney, Sydney, NSW, Australia

[c] Department of Neurosurgery, Brigham and Women's Hospital, Harvard Medical School, Boston, MA, USA

[d] Speech and Hearing Bioscience and Technology, Harvard Medical School, Boston, MA, USA

[e] School of Computer Science and Engineering, University of New South Wales, Sydney, NSW, Australia

[f] Center for Morphometric Analysis, Departments of Psychiatry and Neurology, Athinoula A. Martinos Center for Biomedical Imaging, Massachusetts General Hospital, Harvard Medical School, Boston, MA, USA

[g] Psychiatry Neuroimaging Laboratory, Department of Psychiatry, Brigham and Women's Hospital, Harvard Medical School, Boston, MA, USA

[h] University of Electronic Science and Technology of China, Chengdu, Sichuan, China

Corresponding Authors: Fang Zhang (zhangfanmark@gmail.com) and Lauren J. O'Donnell (odonnell@bwh.harvard.edu)

**Abstract**

We propose a geometric deep-learning-based framework, TractGeoNet, for performing regression using diffusion magnetic resonance imaging (dMRI) tractography and associated pointwise tissue microstructure measurements. By employing a point cloud representation, TractGeoNet can directly utilize tissue microstructure and positional information from all points within a fiber tract without the need to average or bin data along the streamline as traditionally required by dMRI tractometry methods. To improve regression performance, we propose a novel loss function, the Paired-Siamese Regression loss, which encourages the model to focus on accurately predicting the relative differences between regression label scores rather than just their absolute values. In addition, to gain insight into the brain regions that contribute most strongly to the prediction results, we propose a Critical Region Localization algorithm. This algorithm identifies highly predictive anatomical regions within the white matter fiber tracts for the regression task. We evaluate the effectiveness of the proposed method by predicting individual performance on two neuropsychological assessments of language using a dataset of 20 association white matter fiber tracts from 806 subjects from the Human Connectome Project Young Adult dataset. The results demonstrate superior prediction performance of TractGeoNet compared to several popular regression models that have been applied to predict individual cognitive performance based on neuroimaging features. Of the twenty tracts studied, we find that the left arcuate fasciculus tract is the most highly predictive of the two studied language performance assessments. Within each tract, we localize critical regions whose microstructure and point information are highly and consistently predictive of language performance across different subjects and across multiple independently trained models. These critical regions are widespread and distributed across both hemispheres and all cerebral lobes, including areas of the brain considered important for language function such as superior and anterior temporal regions, pars opercularis, and precentral gyrus. Overall, TractGeoNet demonstrates the potential of geometric deep learning to enhance the study of the brain's white matter fiber tracts and to relate their structure to human traits such as language performance.

[1] Abbreviations: AF: arcuate fasciculus, CB: cingulum bundle, EmC: extreme capsule, ILF: inferior longitudinal fasciculus, IOFF: inferior occipitofrontal fasciculus, MdLF: middle longitudinal fasciculus, SLF: superior longitudinal fasciculi, UF: uncinate fasciculus

# 1 Introduction

The brain's white matter connections (fiber tracts) and their tissue microstructure can be quantitatively mapped using diffusion magnetic resonance imaging (dMRI) tractography (F. Zhang et al., 2022). This mapping enables the study of the brain's white matter structural connectivity, a critical substrate for human cognition. Recent investigations in cognitive neuroscience highlight the importance of predicting individual behaviors or traits from individual measures of brain connectivity (Abdallah et al., 2023; Finn and Rosenberg, 2021; Gabrieli et al., 2015; Liu et al., 2023; Shen et al., 2017). By performing individualized predictions of phenotypic measures using methods that can generalize to novel individuals, these approaches can improve our understanding of the organization of the brain (Rosenberg et al., 2018; Scheinost et al., 2019). This prediction is usually achieved by performing a regression task that uses input neuroimaging data to predict an output phenotypic measure, such as cognitive performance. (Regression, which relates a dependent variable to one or more independent (explanatory) variables, is an important technique for data prediction tasks.) Many studies have successfully used dMRI as an input modality to predict phenotype information at the individual subject level (Chen et al., 2020; Dhamala et al., 2020; Feng et al., 2022; Gong et al., 2021; Jeong et al., 2021; Liu et al., 2023; Ooi et al., 2022; Xue et al., 2022). However, these studies used images or summary information from tractography and were unable to benefit from fine-grained, detailed information about individual points within a fiber tract and their tissue microstructure for prediction. In this work, we investigate the prediction of phenotypic information, specifically performance on language assessments, using highly detailed fiber tract information as input. We investigate solutions to several challenges, including the computational representation of white matter fiber tract geometry and microstructure, the design of a deep network that can leverage fiber tract information as input, the improvement of regression performance for predicting language assessment scores, and the interpretation of prediction results in the context of white matter fiber tract and cortical anatomy.

A critical computational challenge in the analysis of white matter fiber tracts is how to represent tracts and their tissue microstructure. dMRI tractography of a single fiber tract, such as the arcuate fasciculus (Fig. 1a), can contain thousands of streamlines that trace the course of the fiber pathway. These streamlines are encoded as sequences of points, such that one tract can contain several hundred thousand points, with associated tissue microstructure information at each point. Many investigations analyze only a single summary statistic per tract (Fig. 1b), such as the number of streamlines (Yeh et al., 2021) or mean fractional anisotropy (Zekelman et al., 2022), ignoring the known spatial variation of tissue microstructure along fiber tracts. More advanced approaches perform tractometry (Fig. 1c) to analyze averages (O'Donnell et al., 2009; Yeatman et al., 2012) or distributions (Chandio et al., 2020) of microstructure measures in subregions along the lengths of fiber tracts. However, due to the need to bin data along tracts, streamline-specific or pointwise information is generally obscured, and it is assumed that the bins correspond across subjects despite the known intersubject variability in shapes, lengths, volumes, and cortical connectivity of tracts (Yeh, 2020). In contrast to these traditional representations, we propose to represent a complete white matter tract with its set of raw points for microstructure analysis using a point cloud, which is an important type of geometric data structure (Fig. 1d). Point cloud representations have previously been utilized to leverage positional information of streamline points (e.g., point coordinates) in tractography data processing tasks such as tractography segmentation and tractogram filtering (Astolfi et al., 2020; Xue et al., 2023). In this study, we investigate the effectiveness of representing a whole white matter tract and its microstructure measurements using a point cloud. In this way, we can directly utilize not only positional information but also tissue microstructure information from all points within a fiber tract and avoid the need for along-tract feature extraction. A point cloud can encode detailed information about the tissue microstructure as well as the three-dimensional shape and spatial extent of fiber tracts. We leverage this point cloud representation for input to a high-level learning task of prediction of neuropsychological scores. We hypothesize that using more detailed white matter fiber tract information can improve the prediction of cognitive measures and enable the localization of critical regions for prediction.

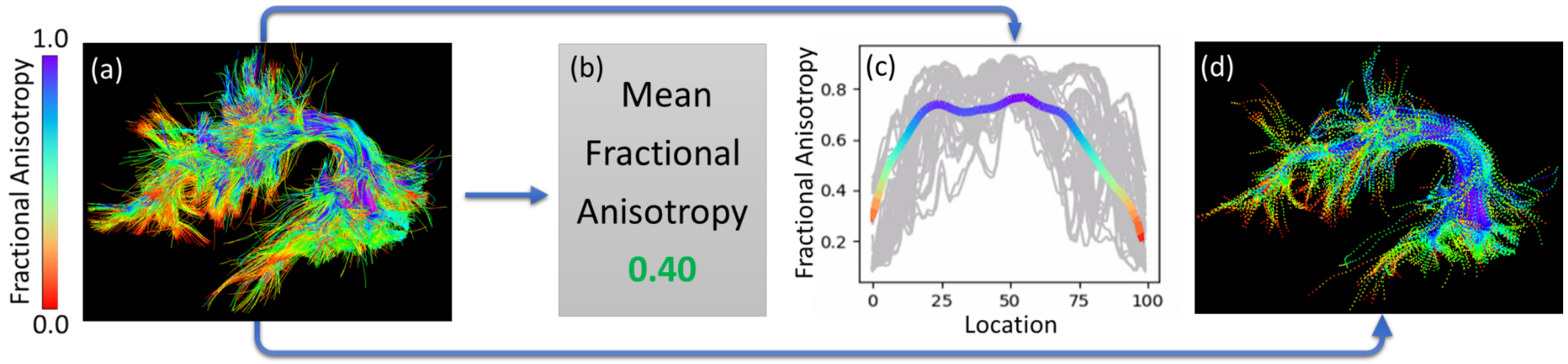

Fig. 1. Different ways to represent a white matter tract and its microstructure measurements. (a) The arcuate fasciculus tract from an example subject with different colors indicating values of fractional anisotropy; (b) Representation with a single mean value; (c) Along-tract representation; (d) Representation with a point cloud.

Designing a high-performance regression model that can handle input geometric data is challenging. Traditional regression algorithms, including linear and non-linear models (such as ElasticNet and Random Forest), have been used in the neuroimaging-based prediction context (Cui and Gong, 2018; Feng et al., 2022; Huang et al., 2016). Deep learning, including multilayer perceptrons (MLP) and convolutional neural networks (CNNs), has also shown promise in predicting cognitive measures (Feng et al., 2022; Jeong et al., 2021; Xue et al., 2022). However, a limitation of such deep learning models is that they can only process input features in the form of raw images or feature vectors and are unable to handle other feature formats that may be more informative, e.g., a point cloud representation of geometric data. Point-based neural networks, such as PointNet, have thus been developed to process geometric data represented as point clouds and demonstrated superior performance (Chen et al., 2017; Qi et al., 2017; Vora et al., 2020). These neural networks are specifically designed to handle irregular and unordered input data, and they have not been used to perform regression based on quantitative representations of tractography geometry. This represents an unexplored opportunity to leverage the strengths of point-based neural networks in this domain.

Another important research challenge is the improvement of performance in regression-based prediction. In particular, we investigate how to leverage the regression label information to enhance training. Unlike a classification task, where the labels are a series of independent discrete values, the labels for a regression task are continuous values with quantitative meaning. While existing regression methods often utilize the predicted label of each subject individually, recent studies have leveraged additional information about the relationship between the predicted labels across subjects to improve prediction performance. For example, the ranking loss (Le Vuong et al., 2021; Liu et al., 2018) utilizes the ranking of the predicted labels across subjects to inform learning. Our recent work has proposed using the differences between regression labels of subjects to define positive or negative pairs in contrastive learning for regression based on tabular data (Xue et al., 2022). However, these works have not used the quantitative label difference information to guide the training of a regression model. Thus, it is worthwhile to explore the development of novel neural networks that can effectively and quantitatively leverage continuous regression scores to improve tract-based neuropsychological score prediction.

Finally, the challenge of the interpretation of neuroimaging-based prediction results has also attracted substantial attention (Chen et al., 2023; Cui et al., 2022; Dan et al., 2022; Feng et al., 2022; Li et al., 2021; Zhang et al., 2018a, 2016; S. Zhang et al., 2022). Interpretation refers to identifying the critical brain regions or features that contribute the most to a prediction task. This interpretation can provide insights into the underlying neural mechanisms related to the predicted variable (Jiang et al., 2022; Kohoutová et al., 2020). Some studies have explored results interpretation in prediction tasks using features from white matter connections (Chen et al., 2023; Feng et al., 2022; Kawahara et al., 2017; Xue et al., 2022). For instance, some methods rank the importance of each input feature, where features are summarized measurements from all points within a white matter connection (Feng et al., 2022; Xue et al., 2022). Another recent study adopts an attention module to identify fiber clusters that predict a specific variable (Chen et al., 2023). However, these methods can only identify entire white matter connections as important for the prediction task and do not allow exploration of the contributions of different regions or individual points within the white matter connections. Therefore, there is a need to develop interpretation methods specifically tailored for prediction tasks based on detailed point cloud representations of white matter fiber tracts.

As an initial testbed for our proposed framework in this paper, we focus on predicting two cognitive measures related to language performance. Previous works have not investigated the machine learning prediction of cognitive measures using input geometric representations of fiber tracts. However, several studies have applied deep learning methods to predict general cognitive (Yeung et al., 2023) and language (Feng et al., 2022) performance using structural connectivity matrices as input. While the connectivity matrix is a powerful abstraction for network analysis of the brain, it includes only scalar connection "strength" information. It thus cannot leverage detailed geometric or microstructure information from tractography. In addition to deep learning methods, traditional statistical methods have shown that measures from individual fiber tracts, such as mean microstructure values or estimates of connectivity strength, significantly relate to neuropsychological measures of language (Ivanova et al., 2021; Liu et al., 2023; Sánchez et al., 2023; Yeatman et al., 2011; Zekelman et al., 2022). By investigating a more detailed fiber tract representation, this investigation can potentially contribute to a deeper understanding of the specific white matter pathways that underlie language abilities.

This study presents a novel geometric deep learning framework, TractGeoNet, for predicting neuropsychological assessment scores and localizing highly predictive regions within white matter tracts. TractGeoNet is designed as a supervised deep-learning pipeline for regression tasks. The paper outlines four key contributions as follows. First, we utilize point cloud representations to preserve the microstructure measurement information from all points within the white matter tracts. This approach provides a comprehensive representation of fiber tract data that can benefit machine learning tasks. Second, we introduce a Paired-Siamese Regression loss for regression to effectively utilize information about the differences between continuous regression labels (language neuropsychological scores in this case). This loss function considers the paired relationship between samples, improving the regression performance. Third, we propose a Critical Region Localization (CRL) algorithm to identify critical regions within each white matter tract. This algorithm enables the localization and interpretation of important regions that contain points that highly contribute to the prediction task. Fourth, we evaluate the proposed TractGeoNet on a large-scale white matter tract dataset of 20 white matter tracts from 806 subjects obtained from the Human Connectome Project (HCP) (Van Essen et al., 2013). The results demonstrate the effectiveness of the proposed approach for predicting neuropsychological scores and identifying critical regions within the tracts.

The current paper extends a preliminary version of the work (Chen et al., 2022) by incorporating several improvements and additional analyses. First, we improve the CRL algorithm by identifying critical regions that exhibit high consistency across multiple trained models and high correspondence across subjects through group-wise analysis. Second, we expand the analysis from one tract (the arcuate fasciculus in the conference publication) to include a total of 20 white matter tracts. Third, we investigate the prediction of an additional language-related neuropsychological assessment and compare the localized critical regions across assessments. By incorporating these improvements, the paper strengthens the overall methodology and provides a more comprehensive evaluation of the proposed TractGeoNet framework.

## 2 Methods

An overall conceptual representation of the framework used to predict neuropsychological scores and localize highly predictive regions based on pointwise analysis of tract microstructure is presented in Fig. 2. In this work, we have focused on two types of brain data included in the HCP dataset, namely dMRI data and neuropsychological data. From the dMRI data, we generated whole brain tractography for each subject, identified each individual's white matter tracts, and extracted their quantitative measures. Then we represented white matter tracts as point clouds and applied the proposed TractGeoNet to predict neuropsychological scores and localize highly predictive regions from the white matter tracts. The following paragraphs of this section describe each of these steps in greater detail.

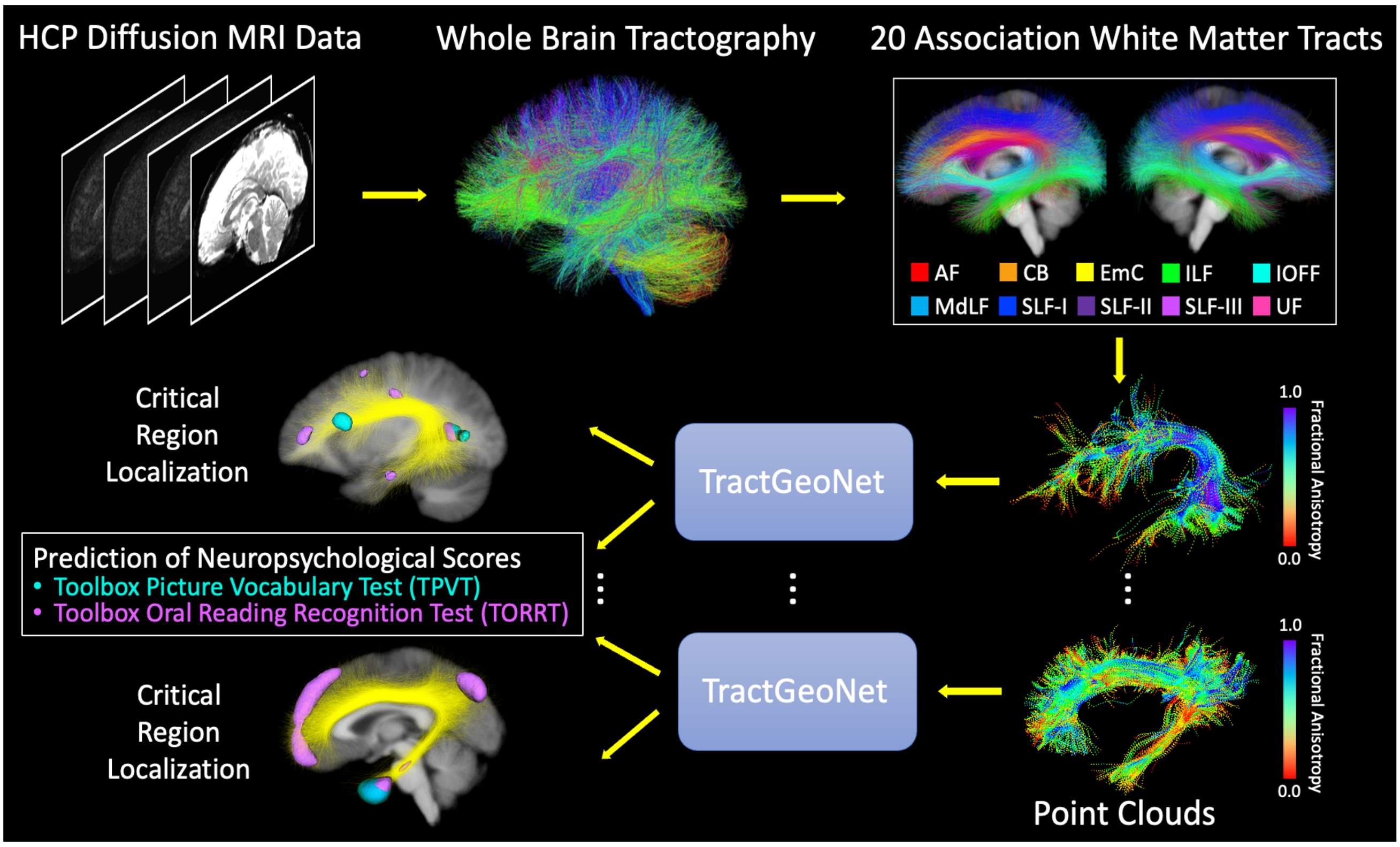

Fig. 2. Schematic of overall methodology. First, dMRI data and neuropsychological assessment scores (TPVT and TORRT) were obtained from the HCP. Whole brain tractography was then computed using individual dMRI data from 809 participants. Next, using the whole brain tractography, we identified 20 association white matter tracts (10 per hemisphere) with their quantitative measurements. Then the white matter tracts with their measurements were represented as point clouds and input to the proposed TractGeoNet. The outputs of TractGeoNet were the predicted neuropsychological assessment scores and localized critical regions that contributed strongly to the prediction results. HCP: Human Connectome Project.

### 2.1 Study Dataset

To evaluate the performance of our proposed method, this study utilized dMRI data and neuropsychological assessment scores from the HCP, a large multimodal dataset acquired from healthy young adults (Van Essen et al., 2013). The dMRI data was collected for a subset of 1065 subjects in the HCP data. From the 1065 subjects, 100 subjects with dMRI data were excluded from this analysis because they were used to develop the ORG white matter atlas employed in this project (described in Section 2.2, see Zhang et al., 2018b for details). Furthermore, to control for the genetic family structure inherent to this dataset, 154 monozygotic twin siblings were excluded due to their nearly identical genetics. Two subjects with incomplete neuropsychological scores were also excluded. The final dataset used in our study thus included 809 unrelated subjects, including 382 male subjects and 427 female subjects, aged from 22 years to 36 years with a mean of 28.6 years.

The investigated dMRI data came from the HCP minimally preprocessed dataset (b = 1000, 2000 and 3000 s/mm2, TE/TR =89/5520 ms, resolution = 1.25 × 1.25 × 1.25 $mm^3$) (Van Essen et al., 2013). For each subject, the dMRI data from the b = 3000 shell of all 90 gradient directions and all 18 b = 0 scans were extracted to perform tractography because this single shell can reduce computation time and memory usage while providing the highest angular resolution for tractography (Descoteaux et al., 2007; Ning et al., 2015; Zekelman et al., 2022).

We investigated language performance using two computerized measures from the NIH Toolbox Cognition Battery, namely the NIH Toolbox Picture Vocabulary Test (TPVT) and the Toolbox Oral Reading Recognition Test (TORRT) (Weintraub et al., 2013). TPVT and TORRT are neurocognitive assessments that capture different aspects of language function. TPVT measures vocabulary comprehension and is a receptive language assessment (Gershon et al., 2014). During this assessment, participants heard a pre-recorded word and were presented with four images; participants were then asked to match the heard word to its associated image (Gershon et al., 2013). TORRT measures reading decoding and is a spoken language assessment (Gershon et al., 2014). During this assessment, participants were presented with a word or letter on a blank screen and were asked to read the presented word or letter aloud (Gershon et al., 2013). Together, TPVT and TORRT are language assessments that have been interpreted to access an individual's "semantic memory," which pertains to general information, facts, and knowledge (Gershon et al., 2013; Renoult et al., 2012; Setton et al., 2022). Assessments were administered in English, and this study utilized the uncorrected scores from each assessment; uncorrected scores are understood to reflect one's absolute level of cognitive capacity compared to the average U.S. individual (Casaletto et al., 2015). In this study cohort, TPVT scores of all investigated subjects ranged from 90.69 to 153.09 with a mean of 117.12, and TORRT scores ranged from 84.20 to 150.71 with a mean of 116.86.

### 2.2 Tractography, White Matter Tract Identification, and Diffusion Measure Extraction

Whole brain tractography was generated from each subject's diffusion MRI data using a two-tensor unscented Kalman filter (UKF) method (Malcolm et al., 2010), as implemented in the ukftractography package (https://github.com/pnlbwh/ukftractography). In contrast to other tractography methods that fit a model to the diffusion signal independently at each voxel, the UKF framework employs prior information from the previous step during each tracking step to help stabilize model fitting (Zhang et al., 2018b). The two-tensor model adopted in the UKF tractography algorithm can depict crossing streamlines, which are prevalent in white matter tracts (Farquharson et al., 2013; Vos et al., 2013). In this way, the first tensor is associated with the tract being traced, while the second tensor captures the streamlines that cross through the tract.

White matter tracts were identified with a robust machine learning approach that has been shown to consistently extract white matter connections across datasets, acquisitions, and the human lifespan (Zhang et al., 2018b). This method was implemented with the WMA package (Zhang et al., 2018b), which uses a well-established fiber clustering pipeline (O'Donnell et al., 2012; O'Donnell and Westin, 2007) in conjunction with an anatomical white matter tract atlas, namely the ORG atlas (Zhang et al., 2018b). In our study, twenty long-range association tracts (n=20) were selected for this investigation due to their investigated relationships to language function (Chang et al., 2015; Forkel and Catani, 2019; Zekelman et al., 2022). The investigated 20 association white matter tracts include bilateral (both left and right hemisphere) arcuate fasciculus (AF), cingulum bundle (CB), extreme capsule (EmC), inferior longitudinal fasciculus (ILF), inferior occipitofrontal fasciculus (IOFF), middle longitudinal fasciculus (MdLF), superior longitudinal fasciculi I, II and III (SLF-I, SLF-II and SLF-III), and uncinate fasciculus (UF), as shown in Fig. 2.

After identification of white matter tracts, microstructure information within tracts was quantified and used for prediction. For each individual tract, two measurements were calculated: the tract-specific fractional anisotropy (FA) and the number of streamlines (NoS). FA indexes how far a tensor is from a sphere to quantify the diffusion anisotropy of the water molecules and NoS is indirectly related to the "strength" of structural connectivity (F. Zhang et al., 2022). These measurements of the white matter tracts have been commonly used in language-related diffusion imaging studies and found to be significantly related to language performance (Chang et al., 2015; Ivanova et al., 2016; Zekelman et al., 2022).

## 2.3 Point Cloud Construction

A white matter tract consists of numerous streamlines, and each streamline consists of many points, resulting in tens of thousands of points within a tract. In the case of the twenty white matter tracts studied, each tract is represented as a point cloud for each individual subject. This point cloud retains all the points within the white matter tract and their quantitative measurements. Each point is characterized by its three spatial coordinates (x, y, z) and two additional measurement dimensions (FA and NoS, as explained in Section 2.2), hence 5 channels for each point. During each training iteration, the input for the neural network (TractGeoNet, see Section 2.4) is formed by randomly sampling N points from the tract point cloud, resulting in an $N \times 5$ dimensional point cloud. We note that, unlike existing studies that usually represent a fiber tract as a fixed subset of fiber points (Garyfallidis et al., 2012; Zhang et al., 2018b), our method represents each fiber tract using a randomly sampled subset of the fiber points in each training iteration. In this way, after a sufficient number of training iterations, the network can consider all the points within the tract (approximately 300,000 per subject on average). This strategy can also be seen as a natural yet effective data augmentation strategy, where repeated random sampling generates many training samples for each input fiber tract. Fig. 2 depicts example point cloud representations of the investigated arcuate fasciculus and cingulum white matter tracts. See Fig. S1 in Supplementary Material 1 for additional streamline and point cloud visualizations of the investigated 20 white matter association tracts. Tracts were visualized in Slicer via the SlicerDMRI project (Norton et al., 2017; Zhang et al., 2020).

## 2.4 Network Architecture

This work proposes a novel deep learning framework, TractGeoNet, that performs a regression task for TPVT and TORRT score prediction. The overall pipeline of TractGeoNet is shown in Fig. 3. During training, the input of TractGeoNet is a pair of point clouds constructed from white matter tracts (see Section 2.3 for details). A Siamese Network (Chopra et al., 2005) that contains two subnetworks (see Section 2.4.1 for details) is adopted to predict the TPVT/TORRT scores of the input pair. The outputs of TractGeoNet are a pair of predicted scores as well as the difference between them. Based on the outputs, a Paired-Siamese Regression loss (see Section 2.4.2 for details) is calculated to utilize information about the differences between continuous regression scores for the benefit of model training. For inference, one subnet of the Siamese Network is retained with one point cloud as input to perform regression (predict a TPVT or TORRT score). During inference, a Critical Region Localization (CRL) algorithm (see Section 2.5 for details) is applied to localize critical regions that highly contribute to the prediction results.

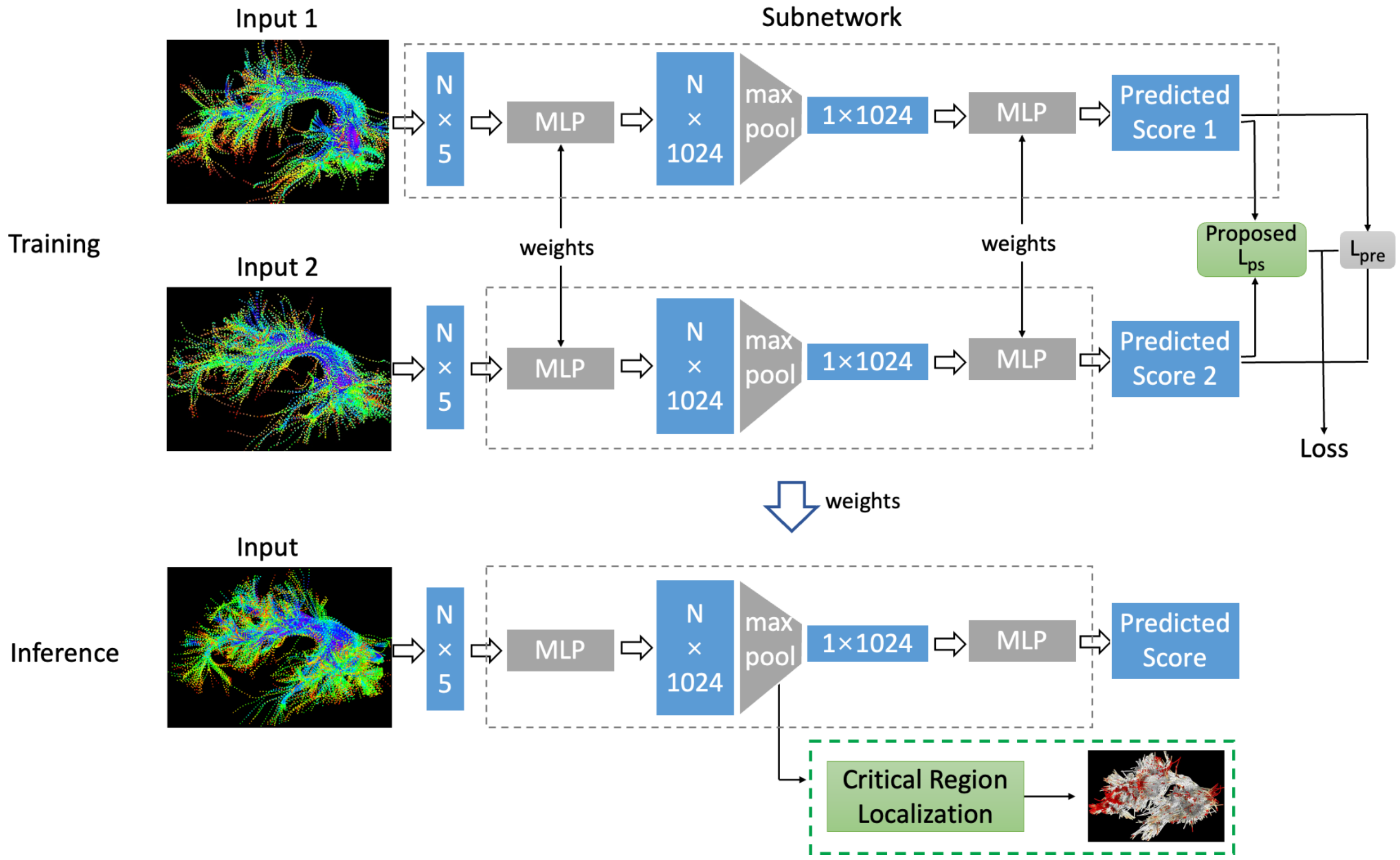

Fig. 3. Overall pipeline of TractGeoNet. A pair of white matter tracts are represented as a pair of point clouds and fed into a point-based Siamese Network. During the training stage, the predicted TPVT or TORRT scores are obtained from the network and a prediction loss ($L_{pre}$) is calculated as the mean of Mean Square Error (MSE) losses between the predicted scores and ground truth scores. The Paired-Siamese Regression loss ($L_{ps}$) is calculated as the MSE between two differences: the difference between predicted assessment scores and the difference between ground truth scores of input pairs. $L_{ps}$ is added to the total loss with weight $w$. During inference, critical regions are recognized with our proposed Critical Region Localization (CRL) algorithm, as shown in the green dashed box.

### 2.4.1 Network Design

In this study, we adopt the Siamese Neural Network (Chopra et al., 2005) as the main architecture of TractGeoNet. To enable the network to leverage the relationships between pairs of inputs, we propose to adopt a Siamese Network that contains two subnetworks with shared weights. A Siamese Network (Chopra et al., 2005) is traditionally adopted to calculate the similarity of two inputs, and here we use it to obtain the difference between predicted scores of an input pair of point clouds (fiber tracts). It includes two subnetworks with shared weights. A subnetwork is used to process an input point cloud and output a prediction score for the corresponding input. The subnetworks of our Siamese Network are developed from PointNet (Qi et al., 2017), which has been widely applied and demonstrated superior performance in applications of point cloud analysis (Aoki et al., 2019; Paigwar et al., 2019; Qi et al., 2017; Xue et al., 2023). As previously described in the Introduction, PointNet is well suited to analyze input unstructured geometric data such as fiber tracts derived from tractography. To perform the regression task, the output dimension of the last linear layer is set to 1 to obtain one predicted score. T-Net (the spatial transformation layer) is removed from PointNet to preserve anatomically important information about the spatial position of tracts (Xue et al., 2023).

### 2.4.2 Proposed Paired-Siamese Regression Loss

The most important aspect of our TractGeoNet's design is the supervision of neural network backpropagation using information from the difference of continuous labels (TPVT and TORRT scores). This leverages the fact that a regression task has continuous scores as ground truth. The intuition behind this method is that individual differences in regression labels (language performance) are informative and can help train the network. Rather than only using the ground truth language performance of each subject for training, we propose a novel loss that considers pairs of subjects and utilizes their predicted and ground truth differences in language assessment scores, aiming at improving regression performance.

We propose the Paired-Siamese Regression loss ($L_{ps}$) to guide the prediction of assessment scores. It considers the relationship between the input pair when training the model. The goal of the loss is to constrain the difference between the predicted scores of the input pair to be the same as the difference between the ground truth scores, thus helping to learn discriminative features for the prediction task. $L_{ps}$ is defined as the mean squared error (MSE) between two differences, where the first is the difference between ground truth scores of an input pair and the second is the difference between predicted assessment scores of an input pair, as follows:

$$L_{ps} = 1/N_b \sum_i ((y_{i1} - y_{i2}) - (\hat{y}_{i1} - \hat{y}_{i2}))^2$$

where $y_{i1}$ and $y_{i2}$ are the labels of the input pair, $\hat{y}_{i1}$ and $\hat{y}_{i2}$ are the predicted scores of the input pair, and $N_b$ is the batch size. In addition to the proposed difference loss, the prediction losses of each input are calculated as the MSE loss between the predicted score and the ground truth score. Then the overall prediction loss of the input pair, $L_{pre}$, is calculated as the mean of the two prediction losses. Therefore, the total loss of our network is $L = L_{pre} + w * L_{ps}$, where $w$ is the weight for $L_{ps}$.

## 2.5 Critical Region Localization Algorithm

In this study, we propose a Critical Region Localization (CRL) algorithm to identify critical regions within fiber tracts that are important in predicting individual language neuropsychological assessment scores during inference. It is integrated into the TractGeoNet pipeline by following the max-pooling layer in the network, as shown in Fig. 3. The algorithm includes two stages, subject-wise contributing point identification and group-wise critical region localization, explained in detail as follows.

The first stage identifies subject-wise points that highly contribute to the prediction task. This stage takes an individual white matter tract as the input to TractGeoNet. We propose to identify contributing points and assign weights to them based on their importance for prediction. To ensure that the whole white matter tract is utilized for interpretation, we divide the tract into multiple sub-point sets (N points per set) through random sampling without replacement and sequentially feed these point sets to the trained TractGeoNet for inference. The CRL algorithm takes advantage of the max-pooling operation in the network as in (Qi et al., 2017) to obtain point sets that contribute to the max-pooled features (referred to as contributing points in our algorithm). However, instead of treating all contributing points equally as in (Qi et al., 2017), our method considers the number of max-pooled features of each contributing point and assigns this number to the point as its weight. This process is repeated M times, such that every point in the white matter tract is fed into the TractGeoNet network M times. The weights of each contributing point are summed across all repetitions. As a result, for each tract from each testing subject, we identify its contributing points along with their corresponding weights.

The second stage performs group-wise analysis to localize critical regions for assessment score prediction across all testing subjects. This analysis aims to produce a heatmap where the value of each voxel corresponds to the number of subjects with a highly contributing point at that voxel location. We leverage the mean

T1-weighted image of the ORG atlas (the coordinate system in which the fiber tracts are represented) as a 3D template volume for heatmap creation. For each subject, the weights of all contributing points (as determined in the first stage above) are mapped into their corresponding voxels in this template space. Because the points along streamlines are sparsely sampled (1.7 mm) in comparison to the voxel size (0.7 mm isotropic), we apply a Gaussian filter (FWHM=10mm) to obtain a more continuous representation of the contributing point weights of each subject. The top 5% of voxels with the highest importance weights are then selected for further analysis and represented as a binary mask image. Finally, the group-wise heatmap is obtained by summing these binary images across all subjects, such that the value in each voxel represents the number of subjects with highly contributing points in that voxel. The heatmap is averaged across ten experiment repetitions (different train/validation/test dataset splits as described in Section 2.6) with ten independently trained TractGeoNet models. Finally, to interpret the most consistently important regions for prediction, we threshold this heatmap at 75% of the total number of testing subjects. This strict threshold identifies voxel locations that are highly consistently important for prediction across 75% or more of subjects. This heatmap level is considered to represent strong consistency across subjects, given the known variability in the localization of language function across subjects (Lu et al., 2021) and across languages (Xu et al., 2017). The anatomical regions corresponding to these voxels are then identified as critical regions for the prediction of language assessment scores. Specifically, we use the groupwise anatomical parcellation provided with the ORG atlas to obtain anatomical labels of the FreeSurfer regions that are intersected by the thresholded heatmap regions.

### 2.6 Implementation Details

For model training and performance evaluation, datasets were split into train/validation/test with the ratio 70%/10%/20%, and we repeated each experiment 10 times with different train/validation/test splits (Xue et al., 2022; S. Zhang et al., 2022). The averaged prediction performance of all testing datasets is reported in the experiment results.

In the training stage, our TractGeoNet model was trained for 500 epochs with a learning rate of 0.001. The batchsize for training was 32 and Admax (Kingma and Ba, 2014) was used for optimization. We used a weight decay with the corresponding coefficient 0.005 to avoid overfitting. We tuned the weight of the Paired-Siamese Regression loss *w* (selected from 0.01, 0.1/3, 0.1, 1/3, 1) on the validation dataset for each tract and each prediction task to obtain the best performance. The final choices of *w* for all tracts and tasks are listed in Table 1. The number of input points N was set to 2048, considering the limitation of GPU memory. The number of iterations M for identifying contributing points in the CRL algorithm was set to 10. All experiments were performed on an NVIDIA RTX 2080 Ti GPU using Pytorch (v1.7.1) (Paszke et al., 2019).

Table 1 Weights of Paired-Siamese Regression loss for all white matter tracts across two prediction tasks.

| Tract | AF left | CB left | EmC left | ILF left | IOFF left | MdLF left | SLF-III left | SLF-II left | SLF-I left | UF left |
|---|---|---|---|---|---|---|---|---|---|---|
| TPVT | 0.033 | 0.033 | 0.1 | 0.1 | 1 | 0.033 | 0.033 | 0.1 | 0.033 | 0.1 |
| TORRT | 0.333 | 1 | 1 | 0.333 | 0.333 | 0.1 | 0.033 | 0.333 | 1 | 0.1 |
| Tract | AF right | CB right | EmC right | ILF right | IOFF right | MdLF right | SLF-III right | SLF-II right | SLF-I right | UF right |
| TPVT | 0.033 | 0.333 | 1 | 0.033 | 0.1 | 0.333 | 0.033 | 0.1 | 0.033 | 0.1 |
| TORRT | 0.01 | 0.1 | 0.333 | 0.333 | 1 | 0.1 | 0.333 | 0.01 | 0.333 | 1 |

# 3 Experiments and Results

## 3.1 Evaluation Metrics

The Pearson correlation coefficient (*r*) was adopted as the evaluation metric to quantify the performance of our proposed method and enable comparisons among approaches. It measures the linear correlation between the predicted and ground truth scores and has been widely applied to evaluate the performance of neurocognitive score prediction (Feng et al., 2022; Gong et al., 2021; Kim et al., 2021; Tian and Zalesky, 2021).

## 3.2 Comparison with baseline methods for the prediction of language assessment scores

We compared our proposed TractGeoNet with several baseline methods for each studied white matter tract across the two language assessments. For all methods, features of FA and NoS were used for prediction. First, three representations of FA measurements of the white matter tract, namely mean FA, along-tract FA, and the proposed pointwise FA (Fig. 1) were compared. For the mean FA representation, classical regression models including Linear Regression (LR), Elastic-Net Regression (ENR) (weight of penalty term 0.01), and Random Forest (RF) were performed to predict TPVT and TORRT scores. These models have been widely applied for predicting cognitive performance based on neuroimaging features (Chen et al., 2020; Cui and Gong, 2018; Feng et al., 2022; Huang et al., 2016). Next, along-tract FA was obtained with the Automated Fiber Quantification (AFQ) algorithm (Yeatman et al., 2012) implemented in Dipy v1.3.0 (Garyfallidis et al., 2014). FA features from 100 locations along the tract were generated from AFQ, and NoS values were concatenated with the FA features to form the input feature vector. For along-tract FA, in addition to the LR, ENR, and RF models, the performance of a 1D-CNN model was also investigated. We adopted the model proposed in a recent study that performs age prediction with microstructure measurements (He et al., 2022). The parameters of all comparison methods were fine-tuned to obtain the best performance. Like TractGeoNet, the comparison methods were also run for ten experiment repetitions with the same data splits as TractGeoNet. For each experiment, an *r* value was obtained from the prediction results of all testing subjects. The mean values and standard deviations of *r* values from the ten experiments were calculated and reported as the prediction performance of each method.

To conduct statistical analysis on the results of prediction performance, the calculated correlation coefficients, which were not normally distributed, were first transformed to *z* scores using Fisher's *r*-to-*z* transformation (Keller et al., 2011; Shen et al., 2015; Tobyne et al., 2018). After that, to compare the performances of different methods, a one-way repeated measures Analysis of Variance (ANOVA) was applied to the *z*-transformed correlation coefficients, followed by post hoc pairwise comparisons using paired t-tests between our proposed method and each comparison method.

The performance of all methods for predicting individual Picture Vocabulary Assessment (TPVT) scores from all twenty input tracts is shown in Table 2. The performance of all methods for predicting individual Oral Reading Assessment (TORRT) scores based on these tracts is shown in Table 3. As shown in Tables 2 and 3, the proposed TractGeoNet has the highest prediction performance, as indicated by the highest *r* values among all comparison methods for all studied white matter tracts across both TPVT and TORRT assessment score prediction tasks. The ANOVA analyses show significant differences in the performance of compared methods for predicting both language assessment scores ($p<0.0001$ in all analyses). In addition, post hoc paired t-tests show that TractGeoNet obtains significantly larger *r* values than all comparison methods for all studied white matter tracts across TPVT and TORRT prediction tasks ($p<0.05$ in all analyses), except for the prediction of the TORRT score with the AFQ+1D CNN method based on the right SLF-I tract. The results confirm the effectiveness of the point cloud representation of white matter tracts and the Paired-Siamese Regression loss in improving the prediction performance of both language-related tasks.

Table 2 Quantitative comparison results of TVPT prediction for 20 association tracts

| Methods | Mean + LR | Mean + ENR | Mean + RF | AFQ + LR | AFQ + ENR | AFQ + RF | AFQ + 1D CNN | TractGeo Net |
|---|---|---|---|---|---|---|---|---|
| Input Features | mean FA, NoS | mean FA, NoS | mean FA, NoS | along-tract FA, NoS | along-tract FA, NoS | along-tract FA, NoS | along-tract FA, NoS | pointwise FA, NoS |
| AF left | 0.151 (0.031) | 0.151 (0.033) | 0.112 (0.048) | 0.241 (0.036) | 0.289 (0.056) | 0.175 (0.072) | 0.163 (0.068) | **0.332 (0.052)** |
| AF right | 0.025 (0.009) | 0.009 (0.084) | 0.067 (0.08) | 0.116 (0.056) | 0.103 (0.065) | 0.11 (0.076) | 0.142 (0.038) | **0.313 (0.063)** |
| CB left | 0.079 (0.067) | 0.086 (0.065) | 0.037 (0.073) | 0.066 (0.063) | 0.135 (0.045) | 0.086 (0.038) | 0.158 (0.066) | **0.258 (0.054)** |
| CB right | 0.149 (0.051) | 0.146 (0.051) | 0.102 (0.054) | 0.134 (0.067) | 0.194 (0.068) | 0.126 (0.072) | 0.141 (0.051) | **0.329 (0.037)** |
| EmC left | 0.062 (0.079) | 0.064 (0.079) | 0.046 (0.084) | 0.051 (0.054) | 0.14 (0.063) | 0.062 (0.069) | 0.132 (0.062) | **0.227 (0.041)** |
| EmC right | 0.045 (0.078) | 0.044 (0.075) | 0.031 (0.051) | 0.034 (0.08) | 0.09 (0.092) | 0.104 (0.068) | 0.170 (0.05) | **0.276 (0.038)** |
| ILF left | 0.254 (0.08) | 0.253 (0.08) | 0.198 (0.077) | 0.104 (0.078) | 0.201 (0.063) | 0.182 (0.102) | 0.094 (0.055) | **0.328 (0.058)** |
| ILF right | 0.203 (0.088) | 0.205 (0.088) | 0.182 (0.093) | 0.137 (0.055) | 0.187 (0.053) | 0.144 (0.074) | 0.136 (0.05) | **0.316 (0.061)** |
| IOFF left | 0.131 (0.066) | 0.131 (0.065) | 0.098 (0.062) | 0.124 (0.064) | 0.099 (0.076) | 0.116 (0.072) | 0.093 (0.089) | **0.266 (0.05)** |
| IOFF right | 0.093 (0.077) | 0.102 (0.077) | -0.015 (0.06) | 0.071 (0.103) | 0.114 (0.095) | 0.124 (0.066) | 0.147 (0.059) | **0.280 (0.033)** |
| MdLF left | 0.116 (0.042) | 0.114 (0.043) | 0.115 (0.067) | 0.104 (0.081) | 0.194 (0.072) | 0.129 (0.061) | 0.201 (0.05) | **0.312 (0.035)** |
| MdLF right | 0.137 (0.069) | 0.136 (0.071) | 0.125 (0.053) | 0.045 (0.067) | 0.117 (0.062) | 0.13 (0.076) | 0.17 (0.051) | **0.327 (0.044)** |
| SLF-III left | 0.119 (0.055) | 0.119 (0.057) | 0.14 (0.065) | 0.054 (0.037) | 0.142 (0.053) | 0.222 (0.054) | 0.228 (0.052) | **0.283 (0.036)** |
| SLF-III right | 0.071 (0.071) | 0.068 (0.07) | 0.018 (0.081) | 0.038 (0.079) | 0.134 (0.068) | 0.154 (0.055) | 0.189 (0.038) | **0.258 (0.039)** |
| SLF-II left | 0.166 (0.053) | 0.17 (0.053) | 0.174 (0.048) | 0.134 (0.05) | 0.205 (0.049) | 0.200 (0.043) | 0.185 (0.056) | **0.299 (0.042)** |
| SLF-II right | 0.147 (0.054) | 0.149 (0.054) | 0.102 (0.068) | 0.099 (0.045) | 0.165 (0.065) | 0.188 (0.097) | 0.218 (0.046) | **0.279 (0.058)** |
| SLF-I left | 0.13 (0.065) | 0.128 (0.063) | 0.088 (0.06) | 0.071 (0.045) | 0.104 (0.066) | 0.124 (0.079) | 0.186 (0.063) | **0.270 (0.056)** |
| SLF-I right | 0.121 (0.054) | 0.122 (0.054) | 0.112 (0.076) | 0.052 (0.036) | 0.14 (0.058) | 0.133 (0.096) | 0.205 (0.056) | **0.273 (0.041)** |
| UF left | 0.139 (0.063) | 0.139 (0.063) | 0.115 (0.075) | 0.046 (0.069) | 0.086 (0.073) | 0.052 (0.08) | 0.139 (0.096) | **0.317 (0.04)** |
| UF right | 0.18 (0.049) | 0.181 (0.05) | 0.141 (0.062) | 0.105 (0.043) | 0.169 (0.036) | 0.141 (0.058) | 0.079 (0.031) | **0.302 (0.051)** |

Table 3 Quantitative comparison results of TORRT prediction for 20 association tracts

| Methods | Mean + LR | Mean + ENR | Mean + RF | AFQ + LR | AFQ + ENR | AFQ + RF | AFQ + 1D CNN | TractGeo Net |
|---|---|---|---|---|---|---|---|---|
| Input Features | mean FA, NoS | mean FA, NoS | mean FA, NoS | along-tract FA, NoS | along-tract FA, NoS | along-tract FA, NoS | along-tract FA, NoS | pointwise FA, NoS |
| AF left | 0.136 (0.068) | 0.138 (0.065) | 0.111 (0.069) | 0.203 (0.092) | 0.215 (0.082) | 0.137 (0.08) | 0.173 (0.08) | **0.356 (0.028)** |
| AF right | 0.001 (0.054) | 0.000 (0.059) | 0.067 (0.074) | 0.077 (0.071) | 0.112 (0.073) | 0.129 (0.075) | 0.172 (0.055) | **0.232 (0.052)** |
| CB left | 0.137 (0.074) | 0.132 (0.076) | 0.052 (0.094) | 0.012 (0.084) | 0.132 (0.086) | 0.122 (0.074) | 0.157 (0.058) | **0.256 (0.047)** |
| CB right | 0.186 (0.053) | 0.185 (0.058) | 0.116 (0.052) | 0.162 (0.079) | 0.235 (0.071) | 0.177 (0.058) | 0.177 (0.071) | **0.294 (0.059)** |
| EmC left | 0.053 (0.084) | 0.052 (0.082) | 0.127 (0.087) | 0.041 (0.066) | 0.057 (0.056) | 0.136 (0.068) | 0.207 (0.059) | **0.278 (0.056)** |
| EmC right | 0.005 (0.058) | -0.012 (0.046) | 0.029 (0.057) | 0.023 (0.055) | 0.129 (0.065) | 0.077 (0.05) | 0.167 (0.058) | **0.265 (0.063)** |
| ILF left | 0.255 (0.076) | 0.254 (0.074) | 0.245 (0.065) | 0.096 (0.081) | 0.175 (0.054) | 0.206 (0.082) | 0.148 (0.047) | **0.331 (0.033)** |
| ILF right | 0.216 (0.051) | 0.216 (0.05) | 0.155 (0.046) | 0.114 (0.062) | 0.205 (0.043) | 0.172 (0.066) | 0.145 (0.068) | **0.272 (0.047)** |
| IOFF left | 0.169 (0.048) | 0.173 (0.048) | 0.181 (0.054) | 0.133 (0.07) | 0.162 (0.077) | 0.144 (0.052) | 0.18 (0.055) | **0.296 (0.067)** |
| IOFF right | 0.118 (0.063) | 0.125 (0.063) | 0.064 (0.057) | 0.129 (0.053) | 0.172 (0.038) | 0.171 (0.061) | 0.205 (0.058) | **0.244 (0.030)** |
| MdLF left | 0.045 (0.048) | 0.044 (0.05) | 0.046 (0.079) | 0.109 (0.055) | 0.154 (0.073) | 0.139 (0.067) | 0.205 (0.075) | **0.324 (0.039)** |
| MdLF right | 0.124 (0.07) | 0.125 (0.068) | 0.09 (0.073) | 0.019 (0.072) | 0.073 (0.067) | 0.145 (0.054) | 0.159 (0.058) | **0.256 (0.055)** |
| SLF-III left | 0.124 (0.055) | 0.124 (0.055) | 0.126 (0.065) | 0.111 (0.055) | 0.209 (0.035) | 0.161 (0.08) | 0.179 (0.053) | **0.285 (0.059)** |
| SLF-III right | 0.131 (0.069) | 0.13 (0.069) | 0.115 (0.048) | 0.029 (0.074) | 0.131 (0.068) | 0.118 (0.068) | 0.19 (0.068) | **0.258 (0.052)** |
| SLF-II left | 0.178 (0.066) | 0.181 (0.064) | 0.235 (0.072) | 0.118 (0.068) | 0.212 (0.058) | 0.226 (0.054) | 0.167 (0.047) | **0.315 (0.054)** |
| SLF-II right | 0.189 (0.042) | 0.192 (0.041) | 0.183 (0.052) | 0.046 (0.058) | 0.135 (0.082) | 0.197 (0.064) | 0.168 (0.073) | **0.266 (0.051)** |
| SLF-I left | 0.159 (0.044) | 0.16 (0.042) | 0.143 (0.059) | 0.065 (0.067) | 0.126 (0.06) | 0.179 (0.072) | 0.201 (0.064) | **0.275 (0.04)** |
| SLF-I right | 0.185 (0.06) | 0.186 (0.063) | 0.151 (0.068) | 0.062 (0.062) | 0.152 (0.069) | 0.189 (0.071) | 0.248 (0.055) | **0.263 (0.061)** |
| UF left | 0.141 (0.06) | 0.143 (0.059) | 0.117 (0.061) | -0.033 (0.08) | 0.042 (0.07) | 0.144 (0.057) | 0.191 (0.077) | **0.279 (0.043)** |
| UF right | 0.157 (0.063) | 0.157 (0.063) | 0.139 (0.054) | 0.082 (0.06) | 0.152 (0.057) | 0.152 (0.051) | 0.071 (0.038) | **0.264 (0.03)** |

By comparing performance across different tracts, we can see that the left AF and left ILF tracts are highly predictive of scores on both assessments. Fig. 4 provides a visualization of the $r$ values of the 20 studied white matter tracts obtained by TractGeoNet for the two prediction tasks. See the video in Supplementary Material 2 for a 3D visualization of the $r$ values of the 20 white matter tracts for the two prediction tasks. As we can see from Fig .4, the $r$ values of the Picture Vocabulary Assessment (TPVT) score prediction are generally similar across both hemispheres, while the $r$ values of the Oral Reading Assessment (TORRT) score prediction are generally higher in the left hemisphere.

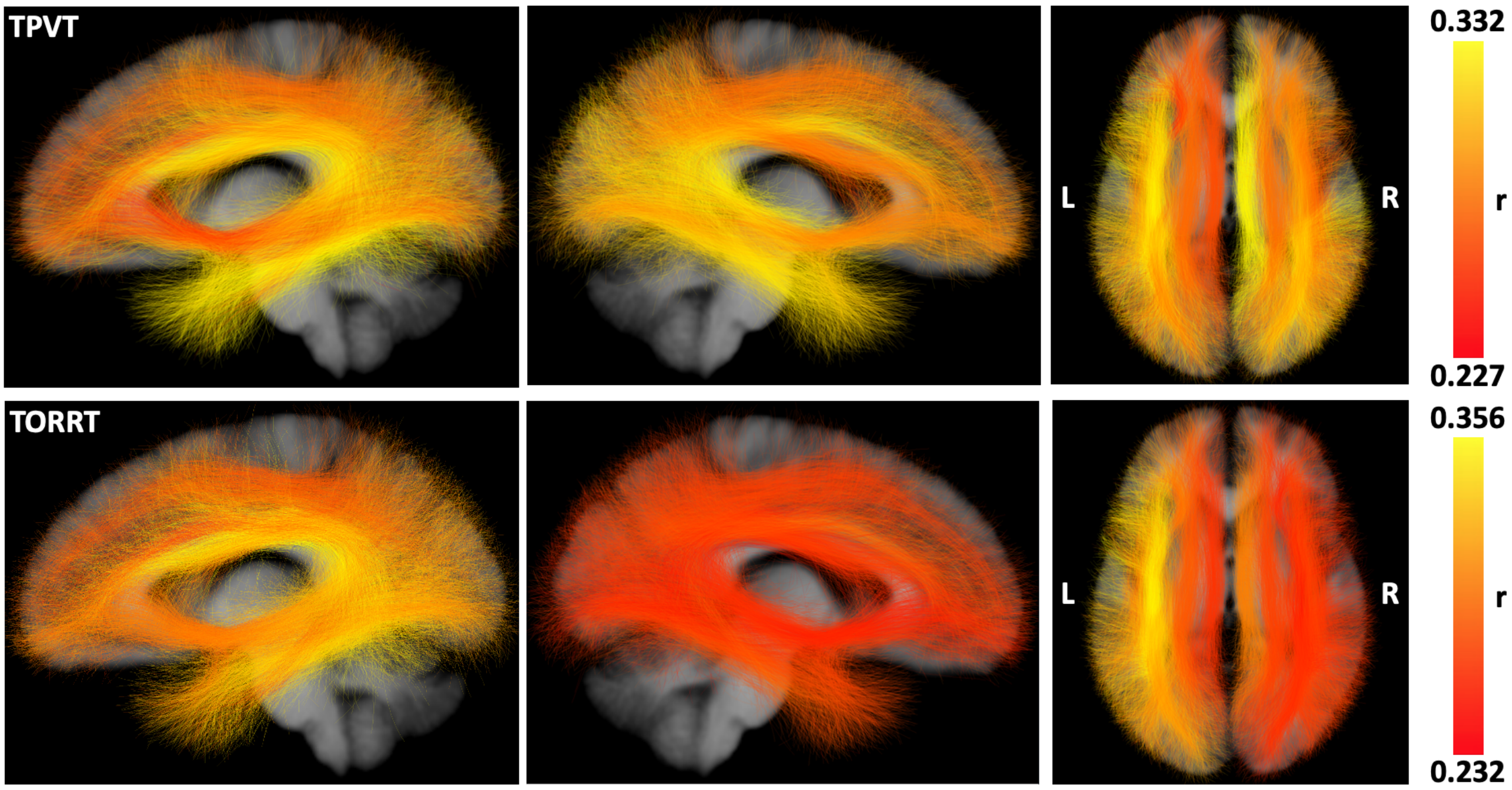

Fig. 4. Visualization of white matter tracts colored according to their $r$ values for the prediction tasks performed by TractGeoNet. The top row shows the results for the prediction of Picture Vocabulary Assessment (TPVT) scores and the bottom row shows the results for the prediction of Oral Reading Assessment (TORRT) scores. The three columns show the white matter tracts from three views (left column: left view; middle column: right view; right column: superior view).

### 3.3 Ablation study

To validate the effectiveness of the proposed Paired-Siamese Regression loss, we performed experiments to investigate the performance of our proposed method with (TractGeoNet) and without (TractGeoNet w/o $L_{ps}$) the Paired-Siamese Regression loss. The $r$ values of all tracts for the two methods and the two prediction tasks are shown in Table 4 and Table 5. As we can see from the table, the proposed novel Paired-Siamese Regression loss consistently improves the performance of our method for all tracts in the prediction of Vocabulary Assessment (TPVT) scores and for almost all tracts in the prediction of Reading Assessment (TORRT) scores, with the exception of the AF right and CB right. This result indicates the effectiveness of utilizing information about the differences between continuous neuropsychological scores to guide the training of the neural network and improve regression performance.

Table 4 Ablation study results of TPVT prediction for 20 association tracts.

| Tract | AF left | CB left | EmC left | ILF left | IOFF left | MdLF left | SLF-III left | SLF-II left | SLF-I left | UF left |
|---|---|---|---|---|---|---|---|---|---|---|
| TractGeo-Net w/o $L_{ps}$ | 0.325 (0.048) | 0.253 (0.034) | 0.221 (0.055) | 0.323 (0.059) | 0.246 (0.064) | 0.299 (0.041) | 0.281 (0.05) | 0.291 (0.043) | 0.255 (0.047) | 0.302 (0.043) |
| TractGeo-Net | **0.332 (0.052)** | **0.258 (0.054)** | **0.227 (0.041)** | **0.328 (0.058)** | **0.266 (0.05)** | **0.312 (0.035)** | **0.283 (0.036)** | **0.299 (0.042)** | **0.270 (0.056)** | **0.317 (0.04)** |
| Tract | AF right | CB right | EmC right | ILF right | IOFF right | MdLF right | SLF-III right | SLF-II right | SLF-I right | UF right |
| TractGeo-Net w/o $L_{ps}$ | 0.305 (0.051) | 0.313 (0.057) | 0.222 (0.061) | 0.300 (0.046) | 0.237 (0.053) | 0.299 (0.066) | 0.250 (0.043) | 0.272 (0.06) | 0.265 (0.043) | 0.279 (0.066) |
| TractGeo-Net | **0.313 (0.063)** | **0.329 (0.037)** | **0.276 (0.038)** | **0.316 (0.061)** | **0.280 (0.033)** | **0.327 (0.044)** | **0.258 (0.039)** | **0.279 (0.058)** | **0.273 (0.041)** | **0.302 (0.051)** |

Table 5 Ablation study results of TORRT prediction for 20 association tracts.

| Tract | AF left | CB left | EmC left | ILF left | IOFF left | MdLF left | SLF-III left | SLF-II left | SLF-I left | UF left |
|---|---|---|---|---|---|---|---|---|---|---|
| TractGeo-Net w/o $L_{ps}$ | 0.345 (0.041) | 0.246 (0.051) | 0.270 (0.078) | 0.321 (0.049) | 0.281 (0.064) | 0.313 (0.045) | 0.266 (0.068) | 0.281 (0.064) | 0.253 (0.052) | 0.268 (0.041) |
| TractGeo-Net | **0.356 (0.028)** | **0.256 (0.047)** | **0.278 (0.056)** | **0.331 (0.033)** | **0.296 (0.067)** | **0.324 (0.039)** | **0.285 (0.059)** | **0.315 (0.054)** | **0.275 (0.04)** | **0.279 (0.043)** |
| Tract | AF right | CB right | EmC right | ILF right | IOFF right | MdLF right | SLF-III right | SLF-II right | SLF-I right | UF right |
| TractGeo-Net w/o $L_{ps}$ | **0.232 (0.050)** | **0.302 (0.038)** | 0.212 (0.04) | 0.258 (0.045) | 0.222 (0.06) | 0.246 (0.052) | 0.241 (0.045) | 0.241 (0.055) | 0.253 (0.067) | 0.235 (0.052) |
| TractGeo-Net | **0.232 (0.052)** | 0.294 (0.059) | **0.265 (0.063)** | **0.272 (0.047)** | **0.244 (0.030)** | **0.256 (0.055)** | **0.258 (0.052)** | **0.266 (0.051)** | **0.263 (0.061)** | **0.264 (0.03)** |

## 3.4 Interpretation of critical regions

### 3.4.1 Subject-wise contributing point identification and group-wise critical region localization

To provide a high-level summary of the CRL algorithm, we give an overview visualization of example results generated from each step of the algorithm, as shown in Fig. 5. First, the identified subject-wise contributing points (with their importance weights) for the prediction of the two investigated language assessments (TPVT and TORRT) are obtained from the neural network directly. Example results from six subjects are shown in Fig. 5 (a). The top 5% of voxels with the highest importance weights are then identified as critical for the prediction task. Next, a group-wise heatmap, where the colors indicate the number of subjects identifying the corresponding voxel as critical, is generated from all testing subjects (as described in Section 2.5), as shown in Fig. 5 (b). This heatmap is averaged across ten experiments with ten independently trained neural network models. Finally, the group-wise critical regions are localized from the group-wise heatmap by setting a threshold (as described in section 2.5), as shown in Fig. 5 (c). Here we choose to visualize the steps of the CRL algorithm using the example of the left AF tract because it shows the highest $r$ values for the prediction of scores on both language assessments. It can be observed that the distributions of contributing points (Fig. 5 (a)) are unique to each individual but show high consistency across different subjects, and they are also consistent with the locations of the identified critical regions in the group-wise heatmap.

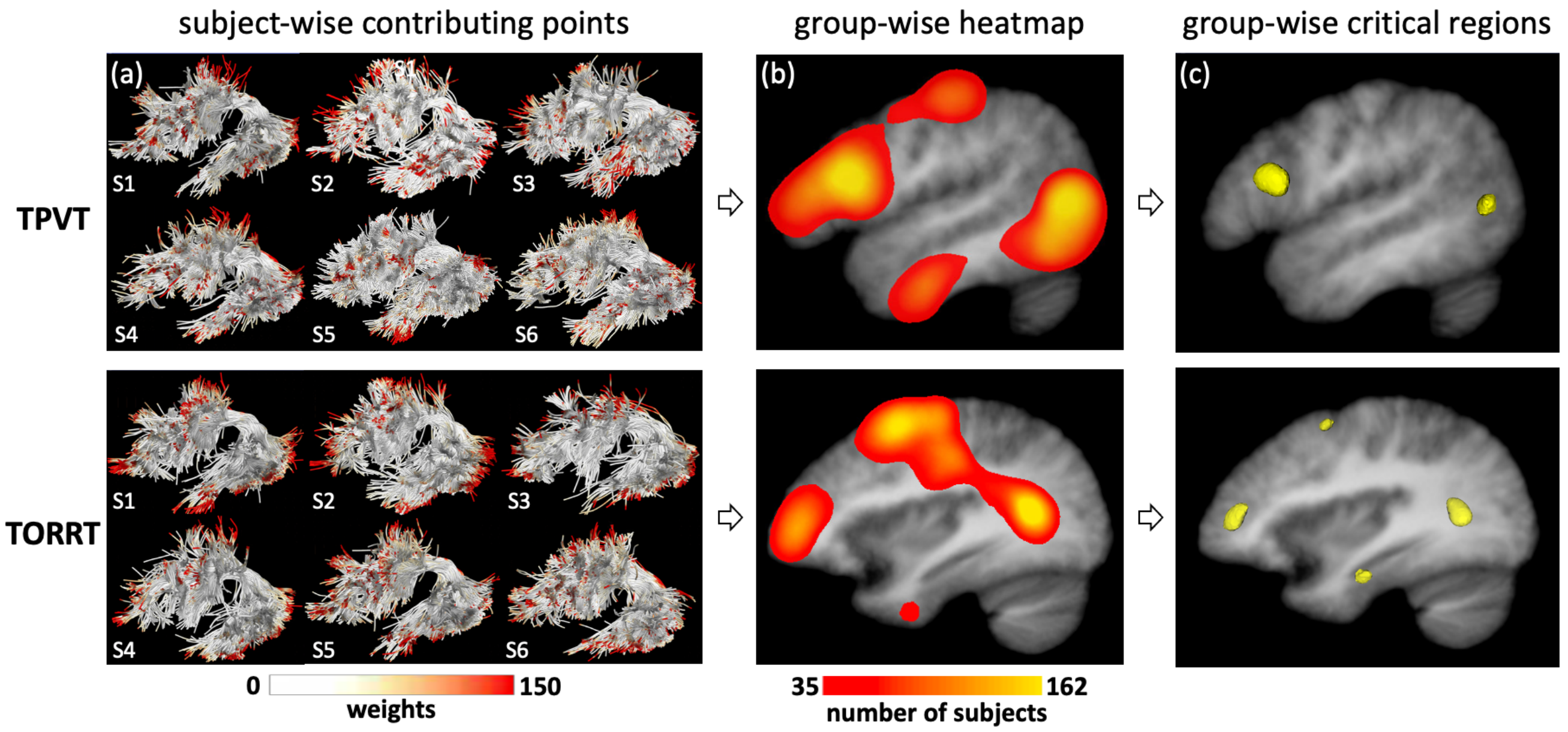

Fig. 5. Visualization of intermediate results generated from each step of critical region localization. The left AF tract is shown as an example here. (a) The identified subject-wise contributing points from six subjects for the two prediction tasks respectively, where color indicates the importance weights of the corresponding points. (b) Group-wise heatmap colored by the number of subjects with highly contributing points in that voxel. (c) Localized group-wise critical regions for language prediction. Top row: Picture Vocabulary Assessment (TPVT). Bottom Row: Oral Reading Assessment (TORRT).

### 3.4.2 Group-wise critical regions for prediction of language assessment scores

Fig. 6 displays localized critical regions highly predictive of Picture Vocabulary Assessment (TPVT) and Oral Reading Assessment (TORRT) scores within 20 white matter tracts. These highly consistent regions are obtained for visualization by thresholding the importance heatmap as described in Section 2.5, using a strict threshold of 75% of subjects. Three-dimensional surface models are shown for each critical region, along with the names of the corresponding FreeSurfer anatomical parcels intersected by the critical region. See the video in Supplementary Material 3 for a 3D visualization of the localized critical regions for both prediction tasks.

As shown in Fig. 6, our method successfully localizes critical regions for predicting scores from two language assessments (TPVT and TORRT) within 20 white matter tracts. It can be observed that multiple highly predictive critical regions can be identified. These regions can be considered to be highly consistent across subjects, as they were highly predictive across multiple experiment repetitions (n=10) and trained models (see details in Section 2.6). While critical regions for prediction were generally identified in tracts from both hemispheres, the distribution of the critical regions is not totally symmetrical across hemispheres. For example, critical regions predictive of TORRT scores were identified in the left AF, but not in the right AF.

To further assess the locations of identified critical regions, we calculated the percentage of critical region voxels located within the cortex and white matter according to the FreeSurfer regions of the voxels. These percentages were averaged across the twenty tracts. For the prediction of TPVT, critical region voxels were located primarily in the cortex (75.4%), while 23.2% were located in the white matter. For the prediction of TORRT, 69.3% of critical region voxels were in the cortex and 28.6% in the white matter. Similarly, for overlapping critical regions (those predictive of scores on both the TPVT and TORRT tasks), 73.1% of voxels were located in the cortex and 24.7% in the white matter.

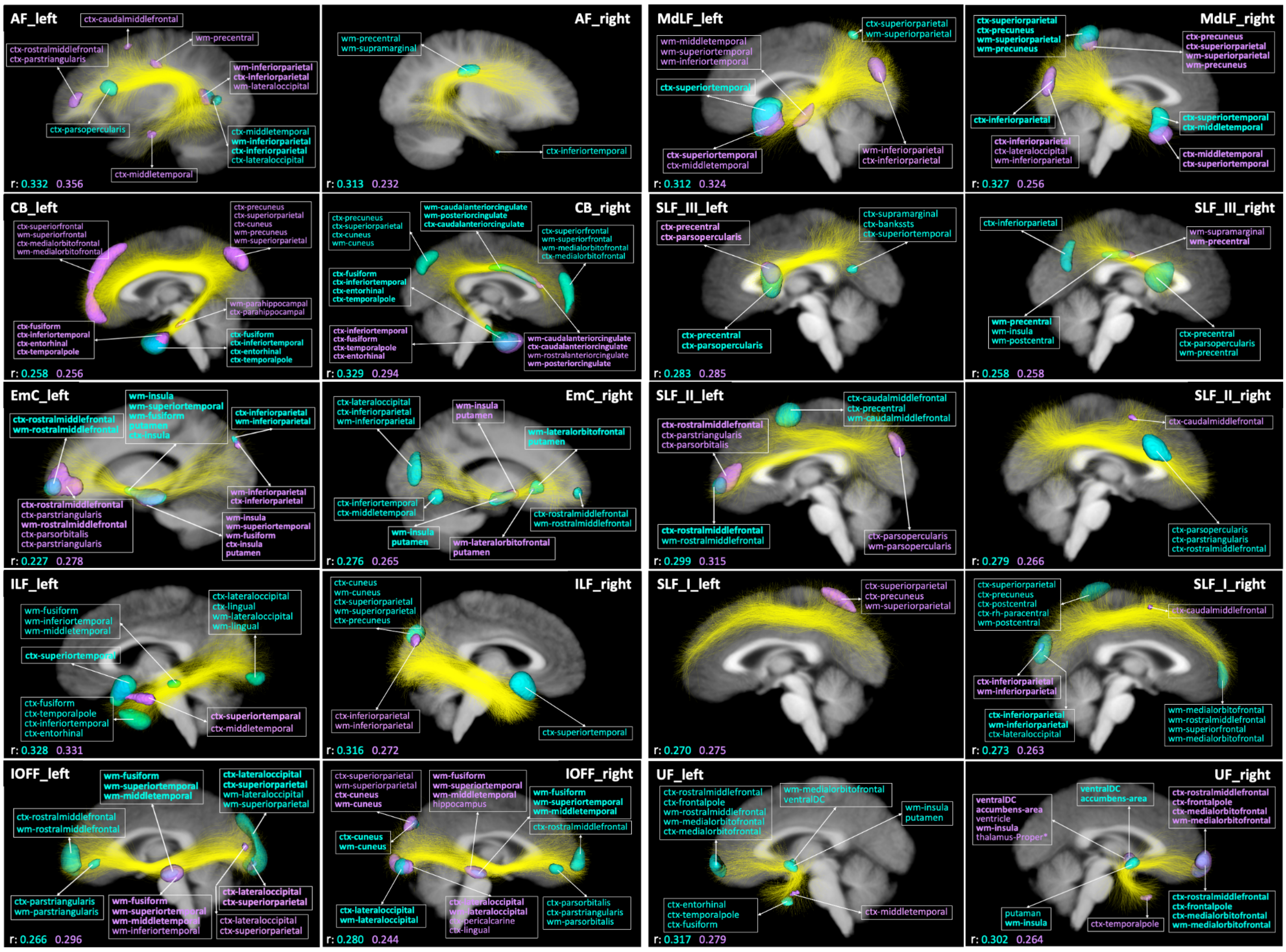

Fig. 6. Visualization of localized critical regions for predicting the Picture Vocabulary Assessment (TPVT, shown in cyan) and the Oral Reading Assessment (TORRT, shown in purple) scores across 20 white matter tracts. Critical regions are labeled with the FreeSurfer anatomical parcels they intersect. Parcels are listed in descending order according to their volume within each critical region, and parcels commonly intersected across both assessments are shown in bold font. The *r* values for the prediction of both assessments are noted in the bottom left corner of each subfigure.

We visualized the overlapping critical regions that were predictive of both TPVT and TORRT scores, as shown in Fig. 7. (To achieve this, for each prediction task we first identified all voxels located in a critical region for at least one tract and created a binary map of these voxels. Then the intersection of the two binary maps provided the overlapping critical regions.) From this visualization, it is apparent that the regions predictive of scores on both assessments are widespread and highly distributed across both hemispheres and all cerebral lobes. The largest regions are located in the frontal, temporal, and parietal lobes, while the smaller regions are located in the occipital lobe and insula. The two largest regions are symmetrically located across hemispheres in the anterior poles of the temporal lobe (regions 1 & 2). Symmetrical regions are also identified in the rostral middle frontal cortex (regions 3 & 10), insular white matter (regions 6 & 11), putamen (regions 6 & 11), and the lateral occipital white matter and cortex (regions 9 & 13). In the left hemisphere, large asymmetrical regions predictive of scores on language assessments are identified in the superior temporal white matter and cortex (regions 4 & 6). Additional asymmetrical left hemisphere regions are located in the precentral cortex and pars opercularis (region 7), and the inferior parietal white matter and cortex (region 14). In the right hemisphere, a large asymmetrical region is located in the superior parietal and precuneus cortex and white matter (region 5).

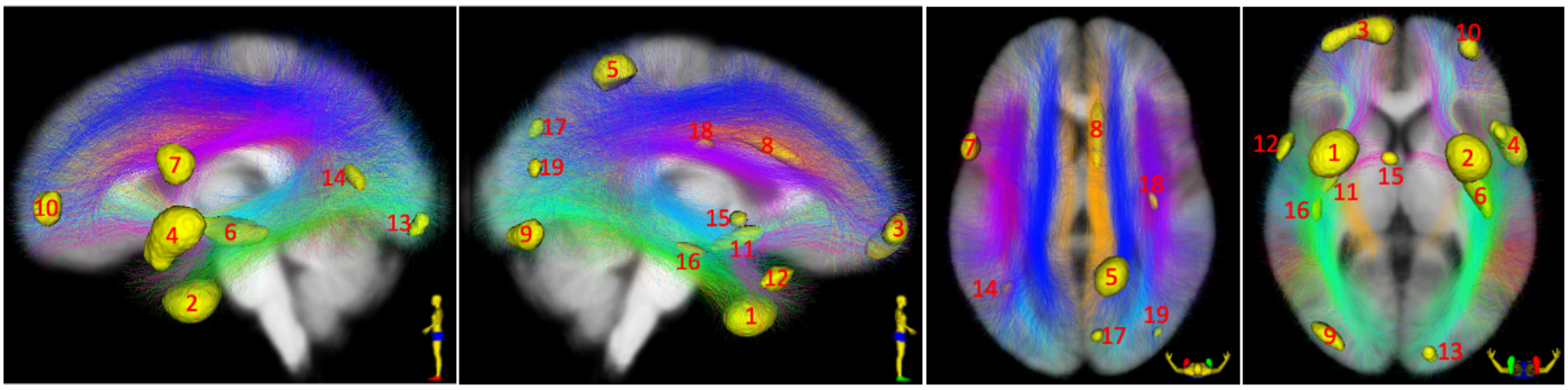

Fig. 7. Summary visualization of critical regions for predicting both assessments. Overlapping critical regions are shown for the prediction of Picture Vocabulary Assessment (TPVT) and Oral Reading Assessment (TORRT) scores across twenty white matter tracts. Regions are shown from four views (from left to right: left, right, superior and inferior view). Regions are labeled according to their sizes with 1 indicating the largest region. The FreeSurfer parcellation labels of the identified regions are listed as follows. 1: ctx-fusiform, ctx-inferiortemporal, ctx-entorhinal, ctx-temporalpole; 2: ctx-fusiform, ctx-inferiortemporal, ctx-entorhinal, ctx-temporalpole; 3: ctx-rostralmiddlefrontal, ctx-frontalpole, ctx/wm-medialorbitofrontal; 4: ctx-superiortemporal; 5: ctx-precuneus, ctx/wm-superiorparietal, wm-precuneus; 6: wm-insula, wm-superiortemporal, wm-fusiform, ctx-insula, putamen; 7: ctx-precentral; ctx-parsopercularis; 8: wm/ctx-caudalanteriorcingulate, wm-precentral; 9: ctx/wm-lateraloccipital; 10: ctx/wm-rostralmiddlefrontal; 11: wm-insula, putamen; 12: ctx-middletemporal, ctx-superiortemporal; 13: ctx/wm-lateraloccipital, ctx-lingual; 14: wm/ctx-inferiorparietal; 15: VentralDC, Accumbens-area; 16: wm-fusiform, wm-superiortemporal, wm-middletemporal; 17: ctx/wm-cuneus; 18: wm-precentral, wm-insula; 19: ctx/wm-inferiorparietal. L: left; R: right; S: superior; I: inferior.

## 4 Discussion

In this paper, we proposed a novel deep learning framework, TractGeoNet, for predicting neuropsychological assessment scores and localizing highly predictive regions within white matter tracts. We utilized a point cloud representation, which preserved pointwise information within the white matter tracts, to benefit machine learning tasks. We introduced a Paired-Siamese Regression loss to effectively utilize information about the differences between continuous regression labels (neuropsychological scores in this case), which improved the regression performance. In addition, we proposed a Critical Region Localization (CRL) algorithm to identify critical regions within each white matter tract, enabling the localization of highly predictive regions for prediction tasks. Our experimental results on a large-scale dataset of 20 white matter tracts demonstrated the effectiveness of the proposed approach for predicting neuropsychological scores and identifying critical regions within the tracts. Several detailed observations about the experimental results are discussed below.

In this study, for the first time, we represented white matter tracts and their pointwise microstructure measurements as point clouds. This representation preserves pointwise information, which benefits prediction performance and enables critical region localization by jointly utilizing point-based neural networks. On the one hand, TractGeoNet with the point cloud representation outperforms all other combinations of input representations and regression models for the 20 white matter tracts, demonstrating its effectiveness in improving prediction performance. On the other hand, the point cloud representation together with point-based neural networks successfully localizes critical predictive regions for language processes. The interpretation of multivariate pattern-based predictive models is a known open challenge in neuroimaging (Kohoutová et al., 2020; Yan et al., 2022). Our proposed method for interpretation via critical region localization extends techniques in the geometric deep learning literature that directly identify points that are highly contributing to learning tasks (Qi et al., 2017). In contrast to existing methods for connectome-based interpretation (Cui et al., 2022; Xue et al., 2022), our proposed method does not require additional feature selection steps or the addition of subnetworks to aid interpretation. One potential strength of our method is that it enables the interpretation of predictive points and regions at both the single-subject and the group level. In contrast to recent publications that often interpret one best-performing model (Adebayo et al., 2018; Chen et al., 2023; Li et al., 2021), in this work, we have performed interpretation using ten independent experiments (ten independently trained models) to avoid bias towards any one particular model or training cohort and improve the robustness of interpretation results. To focus on highly consistent results, we have visualized only the regions that were predictive in over 75% of testing subjects across all models. Several prior studies have also investigated the contributions of different regions within white matter tracts to language processing. The contribution of different subregions of arcuate fasciculus tracts to language performance has been previously studied by using anatomical segments such as anterior and posterior or temporal and frontal (Gao et al., 2022; Ivanova et al., 2021; Wang et al., 2021) or high FA subregions (Yeatman et al., 2011). In contrast to these methods, the proposed critical region localization approach can leverage detailed information to interpret which individual points within a fiber tract are important for prediction.

In this work, the proposed Paired-Siamese Regression loss used quantitative label difference information to guide the training of the regression model and further improved regression performance. Other works have pursued different strategies to improve regression performance by using the relationships between input data pairs. For example, the ranking loss uses order information to inform learned feature representations (Le Vuong et al., 2021; Liu et al., 2018), while our contrastive regression strategy forms positive and negative input data pairs using label information (Xue et al., 2023). We note that a direct comparison of $r$ values to existing work is not straightforward, as we have a unique input data representation. However, one recent study predicted language performance based on the whole brain structural connectome and achieved $r$ values from 0.2 to 0.3 (Feng et al., 2022), which are similar to the results of our method that only utilizes an individual white matter tract.

The methods developed here create new avenues to identify white matter fiber tracts most predictive of scores on neuropsychological assessments and provide the potential to examine their hemispheric differences. Across all investigated association white matter tracts, the left AF is the most predictive of language assessment scores (see Tables 2 and 3). The left ILF is also highly predictive (third most predictive tract for TPVT and second most predictive for TORRT). These findings align with previous language research that consistently links the left AF and ILF to language processes (Chang et al., 2015; Ivanova et al., 2021; Shin et al., 2019; Zekelman et al., 2022). In this study, we also find that the prediction performance ($r$ values) of white matter tracts for Picture Vocabulary Assessment (TPVT) are similar across both hemispheres, while the Oral Reading Assessment (TORRT) prediction performance of white matter tracts is higher in the left hemisphere (see Fig. 4). This result is complementary to investigations that have found effects of laterality may depend upon the particular aspect of language function being probed (e.g., word reading vs. picture vocabulary) (Bradshaw et al., 2017; COLA consortium et al., 2022; Peelle, 2012; Ruff et al., 2008).

Although the Picture Vocabulary Assessment and the Oral Reading Assessments are primarily assessments of language function, it is understood that such assessments are not independent of other cognitive and sensory domains (e.g., auditory, visual, attention, memory, processing speed, etc.) (Harvey, 2019). In this study, the right hemisphere cingulum bundle was found to be the second most predictive tract for TPVT and the anterior cingulum was the most predictive region (Fig. 6). We note that this finding contrasts with a recent study using the same HCP Young Adult dataset, which did not find a strong relationship between cingulum bundle mean FA and language function (Zekelman et al., 2022). However, other groups have found that the FA of the anterior cingulate is highly and uniquely predictive of multiple cognitive functions (including language) (Bathelt et al., 2019). Overall, this finding highlights a potential benefit of pointwise analysis of fiber tracts to identify predictive relationships that may be missed when analyzing tract mean FA, and it suggests the potential of TractGeoNet to identify tracts and regions that may be important to additional functions beyond language.

In addition, the methods developed in this paper create new avenues to visualize and explore critical regions within fiber tracts that predict scores on neuropsychological assessments. In this investigation, critical regions predictive of scores on the Picture Vocabulary Assessment (TPVT) are both unique and overlapping with the critical regions predictive of scores on the Oral Reading Assessment (TORRT). While critical regions unique to each assessment may highlight differences between assessments, overlapping regions may represent "core" regions of the language network. Here we briefly discuss some of the critical regions that are predictive of both TPVT and TORRT scores in relation to previously identified language regions (Fig. 7). For example, the present study identified critical regions within the superior temporal lobe of the left hemisphere (Fig. 7, regions 4 and 6). Similarly, previous fMRI investigations have consistently found increased blood oxygen level-dependent signals spanning the superior temporal gyrus during various language and speech-based tasks (Agarwal et al., 2019; Binder et al., 2000; Capek et al., 2010; Hodgson et al., 2021; Ramos Nuñez et al., 2020). Additionally, we identified large critical regions within the anterior temporal lobe across both hemispheres (Fig. 7, regions 1 and 2). Previous investigations that have used brain stimulation techniques have found bilateral regions in the anterior temporal lobe contribute to semantic representations, an important component of language processes (Lambon Ralph et al., 2009; Shimotake et al., 2015). Finally, this study identified a critical region in the left pars opercularis and left precentral cortex (Fig. 7, region 7). The left pars opercularis and left precentral cortex have been found to be important for speech production; Broca's area is classically defined as a region including the pars opercularis (Dronkers et al., 2007) and damage to the left precentral gyrus has resulted in Broca's aphasia and apraxia of speech (Itabashi et al., 2016; Mori et al., 1989). On average, 26% of the critical regions were located in the white matter and 73% in the cortex (representing the tract origins and terminations in the region of the gray-white matter interface). This shows that the long-range white matter tracts and the distant cortical areas that they connect are both important contributors to the prediction of language function. This largely agrees with the long-held historical understanding that language function depends not only on the brain's cortical surface but also on its underlying connections (Hagoort, 2014; Wernicke, 1874). Our finding also relates to recent work showing that the area of pathway innervation of association tracts, e.g. the region of their cortical origins and terminations, has a large degree of inter-individual variability and is therefore a good descriptor of inter-individual differences in white matter structure (Yeh, 2020).

Finally, we note some limitations of this study and directions for future research. Though many recent investigations predict individual behaviors or traits from individual measures of brain connectivity, this is known to pose an extremely challenging problem, in part because individual performance may be influenced by a plethora of cultural, environmental, and biological factors (Fernández and Abe, 2018; Howieson, 2019; Liu et al., 2020). In this study, we have investigated two neuropsychological assessments that probe language function. Future work can investigate applications of the TractGeoNet framework for additional datasets and additional regression tasks. The TractGeoNeo framework may have future applications as an exploratory tool for hypothesis generation, or to identify informative regions within tracts for downstream statistical analyses. Finally, by representing white matter tracts as point clouds, features are extracted individually from each point. Future technical improvements, such as novel neural network design, could be investigated to incorporate additional information from white matter fiber tract input data such as relationships between neighboring points or information about continuous streamlines. In addition to microstructure data, complementary measurements could be investigated within the TractGeoNet framework, such as local tract shape measures or data from other modalities.

## 5 Conclusion

We have proposed a new framework, TractGeoNet, to enable the detailed analysis of white matter fiber tracts including thousands of points and accompanying microstructure information. The framework can utilize pointwise information by representing white matter tracts as point clouds. We demonstrated that this approach can better predict performance on neuropsychological assessments of language performance in comparison with popular regression methods and common white matter tract representations. Of the twenty tracts studied, we found that the left arcuate fasciculus tract was the most highly predictive of the two studied language performance assessments. Within each tract, we localized critical regions whose microstructure and point information were highly consistently predictive of language performance across different subjects and multiple independent experiments. These critical regions were widespread and distributed across both hemispheres and all cerebral lobes, including known regions important for language such as Broca's area. Overall, this study demonstrates the potential of geometric deep learning to enhance the study of the brain's white matter fiber tracts and their relationship to language and other cognitive functions.

### Declaration of Generative AI and AI-assisted technologies in the writing process

During the preparation of this work the authors used ChatGPT in order to improve the readability of several paragraphs in the Introduction. After using this tool/service, the authors reviewed and edited the content as needed and took full responsibility for the content of the publication.

### Data and code availability

The data used in this project is from the public Human Connectome Project (HCP) dataset. It can be downloaded through the ConnectomeDB (db.humanconnectome.org) website. The code for data analysis will be available upon publication at https://github.com/SlicerDMRI/TractGeoNet.

### Author contributions

Yuqian Chen: Conceptualization, Methodology, Software, Writing - Original draft preparation. Leo R. Zekelman: Result analysis, Writing - Draft and Editing. Chaoyi Zhang: Methodology, Writing- Reviewing and Editing. Tengfei Xue: Methodology, Writing- Reviewing and Editing. Yang Song: Writing- Reviewing and Editing. Nikos Makris: Conceptualization, Writing- Reviewing and Editing. Yogesh Rathi: Writing- Reviewing and Editing. Alexandra J. Golby: Conceptualization, Writing- Reviewing and Editing. Weidong Cai: Resources, Supervision, Writing- Reviewing and Editing. Fan Zhang: Conceptualization, Methodology, Writing - Review & Editing. Lauren J. O'Donnell: Conceptualization, Methodology, Writing - Review & Editing, Supervision, Funding acquisition.

### Acknowledgments

This work was supported by the National Institutes of Health (NIH) grants: R01MH125860, R01MH119222, R01MH132610, R01MH112748, R01NS125307, R01AG042512, K24MH116366, P41EB028741, and R01NS125781.